\documentclass[conference,onecolumn]{IEEEtran}
\usepackage{ifthen}
\newboolean{isanonymous}
\setboolean{isanonymous}{false} % Default: not anonymous

\makeatletter
\@for\opt:=\@classoptionslist\do{%
    \ifthenelse{\equal{\opt}{anonymous}}{\setboolean{isanonymous}{true}}{}
}
\makeatother

\newcommand{\blind}[1]{%
    \ifbool{isanonymous}{\textit{Blinded for review}}{#1}%
}

\IEEEoverridecommandlockouts
\usepackage{cite}
\usepackage{amsmath,amssymb,amsfonts}
\usepackage{graphicx}
\usepackage{textcomp}
\usepackage[table]{xcolor}
\def\BibTeX{{\rm B\kern-.05em{\sc i\kern-.025em b}\kern-.08em
    T\kern-.1667em\lower.7ex\hbox{E}\kern-.125emX}}

\usepackage{algorithm}
\usepackage{algpseudocodex}
\usepackage{glossaries}
\newglossary[ntn]{notation}{not}{ntn}{Notation}
\newacronym{genai}{GenAI}{generative artificial intelligence}
\newacronym{llm}{LLM}{large language model}
\newacronym{epf}{EPF}{electricity price forecasting}
\newacronym{daa}{DAA}{day-ahead auction}
\newacronym{ml}{ML}{machine learning}
\newacronym{rmse}{RMSE}{root mean square error}
\newacronym{mae}{MAE}{mean absolute Error}
\newacronym{smape}{SMAPE}{symmetric mean absolute percentage error}
\newacronym{tsfm}{TSFM}{time series foundation model}
\newacronym{dm}{DM}{Diebold-Mariano}
\newacronym{dst}{DST}{daylight saving time}
\newacronym{cv}{CV}{cross-validation}
\newacronym{api}{API}{application programming interface}
\newacronym{dl}{DL}{deep learning}

\makeglossaries

\usepackage{eurosym}
\usepackage{xurl}

\usepackage{fancyhdr}%for copyright footer
\usepackage{helvet}       % Helvetica font support

\usepackage{orcidlink}

\usepackage{comment}

\usepackage{booktabs}  % For professional-looking tables
\usepackage{multirow}
\usepackage{booktabs}
\usepackage{comment}
\usepackage{xurl}
\usepackage{eurosym}
\usepackage{threeparttable}

\begin{document}
% \title{Bare Demo of IEEEtran.cls\\ for IEEE Conferences}
\title{Benchmarking Pre-Trained Time Series Models for Electricity Price Forecasting
\thanks{
%\vspace{0em}
\noindent\hspace*{-.3cm}\hrulefill
%\rule{5cm}{0.4pt}  % 3cm long, 0.4pt thick line
%\vspace{0.5em}

\blind{This research was funded in part by the Luxembourg National Research Fund (FNR) and PayPal, PEARL grant reference 13342933/Gilbert Fridgen. For the purpose of open access, and in fulfillment of the obligations arising from the grant agreement, the author has applied a Creative Commons Attribution 4.0 International (CC BY 4.0) license to any Author Accepted Manuscript version arising from this submission. The research was carried out as part of a partnership with the energy retailer Enovos Luxembourg S.A}.

© 2025 IEEE. Personal use of this material is permitted. Permission from IEEE must be obtained for all other uses, in any current or future media, including reprinting/republishing this material for advertising or promotional purposes, creating new collective works, for resale or redistribution to servers or lists, or reuse of any copyrighted component of this work in other works.
}
}


\author{\blind{\IEEEauthorblockN{Timothée Hornek\orcidlink{0000-0001-6676-6541}\IEEEauthorrefmark{1},
Amir Sartipi\orcidlink{0000-0002-0124-9823}\IEEEauthorrefmark{1},
Igor Tchappi\orcidlink{0000-0001-5437-1817}\IEEEauthorrefmark{1}, and 
Gilbert Fridgen\orcidlink{0000-0001-7037-4807}\IEEEauthorrefmark{1}}

\IEEEauthorblockA{\IEEEauthorrefmark{1}SnT - Interdisciplinary Center for Security, Reliability and Trust\\
University of Luxembourg\\
Kirchberg, Luxembourg\\
Email: \{timothee.hornek, amir.sartipi, igor.tchappi, gilbert.fridgen\}@uni.lu}}
}
\fancypagestyle{firstpage}{
  \fancyhf{}
  \renewcommand{\headrulewidth}{0pt} % remove header line
  %\fancyfoot[L]{\fontsize{9}{11}\fontfamily{phv}\selectfont\textbf{979-8-3315-1278-1/25/\$31.00 ©2025 IEEE}}
}
\setlength{\footskip}{1.27cm}

% \IEEEoverridecommandlockouts
% \IEEEpubid{\makebox[\columnwidth]{978-1-7281-1234-5/20/\$31.00~\copyright~2025 IEEE \hfill} \hspace{\columnsep}\makebox[\columnwidth]{ }}

\maketitle
\thispagestyle{firstpage} % apply footer only to the first page

\begin{abstract}
%The abstract goes here.
Accurate \gls{epf} is crucial for effective decision-making in power trading on the spot market.
While recent advances in \gls{genai} and pre-trained \glspl{llm} have inspired the development of numerous \glspl{tsfm} for time series forecasting, their effectiveness in \gls{epf} remains uncertain.

To address this gap, we benchmark several state-of-the-art pretrained models—Chronos-Bolt, Chronos-T5, TimesFM, Moirai, Time-MoE, and TimeGPT—against established statistical and \gls{ml} methods for \gls{epf}. Using 2024 \gls{daa} electricity prices from Germany, France, the Netherlands, Austria, and Belgium, we generate daily forecasts with a one-day horizon.

Chronos-Bolt and Time-MoE emerge as the strongest among the \glspl{tsfm}, performing on par with traditional models.
However, the biseasonal MSTL model, which captures daily and weekly seasonality, stands out for its consistent performance across countries and evaluation metrics, with no \gls{tsfm} statistically outperforming it.

\end{abstract}

\glsresetall

% \fi
%\IEEEpeerreviewmaketitle
%\maketitle

\section{Introduction}\label{sec:introduction}
Recent advances in \gls{genai}, particularly the widespread adoption of pre-trained \glspl{llm}, have catalyzed the development of analogous models tailored specifically for time series data, commonly known as \glspl{tsfm}.
Developers describe these models as ``universal forecasters'', capable of providing robust predictions in various domains~\cite{woo_unified_2024}.
Their training on extensive datasets enables their application in various contexts without the need for additional training, relying solely on their inference capabilities.
A specific field of application is \gls{epf}, a well-established time series forecasting task focused on wholesale electricity prices~\cite{weron_electricity_2014}.
In Europe, for example, the primary electricity auction, namely the \gls{daa}, takes place daily at noon, determining a single clearing price for each hour of the following day~\cite{european_commission_consolidated_2021}.

Researchers are currently discussing several \glspl{tsfm} in the literature. 
For example, Chronos models are built on preexisting \gls{llm} architectures, using the T5 model architecture, and trained on a combination of existing and synthetic time series data~\cite{ansari_chronos_2024}. 
Similarly, TimesFM is trained on a large corpus of time series data, employing a decoder-only architecture \cite{das_decoder-only_2024}. 
Time-MoE uses a sparse mixture-of-experts (MoE) architecture, which activates only a subset of its parameters during inference, preserving computational efficiency despite the high total number of parameters.
The model is trained on an extensive dataset that contains 300 billion time points~\cite{shi_time-moe_2024}. 
Morai was trained on the LOTSA dataset, containing more than 27 billion observations, to train its masked encoder architecture~\cite{woo_unified_2024}. 
The weights of the models mentioned so far are publicly available. 
In contrast, TimeGPT is a closed source pre-trained model designed to emphasize simplicity and ease of use by enabling predictions through \gls{api} calls \cite{garza_timegpt-1_2024}.
All these models provide zero-shot forecasting (i.e., forecasting without fine-tuning or additional training)  capabilities, enabling them to use their inference abilities to generate forecasts without requiring any additional prior training.

Forecasting with \glspl{tsfm} remains an active area of research across various fields, including finance and the energy domain. 
For instance, the authors in~\cite{fu_financial_2024} fine-tuned TimesFM~\cite{das_decoder-only_2024} for financial time series forecasting, outperforming several statistical and \gls{dl} benchmark models.
In the energy domain, the authors in~\cite{meyer_benchmarking_2024} trained transformer-based models from scratch and compared their performance with \glspl{tsfm} for household load forecasting.
They found that TimesFM~\cite{das_decoder-only_2024} outperformed the models trained from scratch, particularly when increasing the input size. 
In another study focusing on household power consumption data~\cite{sartipi_bridging_2025}, the authors benchmarked statistical and \gls{ml} models against \glspl{tsfm} for data imputation tasks. 
Their findings suggest that \glspl{tsfm} can improve imputation performance.
Focusing on \gls{epf}, the authors in~\cite{lu_large_2024-1} investigated the integration of market sentiment and bidding behavior into price forecasting models. 
By fine-tuning \glspl{llm} and using conditional generative adversarial networks (CTSGAN), they achieved improved performance in forecasting price spikes. 
Similarly, focusing on \gls{epf}, the authors in~\cite{andrei_energy_2024} categorized time series forecasting approaches into four groups (econometric, deep learning, transformer-based, and \glspl{llm}) and conducted a benchmarking study. 
Their benchmark emphasized \gls{dl} models using transformer-based architectures, testing the models on electricity price data from several European countries.
The results indicated that transformer-based models achieved superior performance in forecasting electricity prices, with the \gls{tsfm} TimesFM~\cite{das_decoder-only_2024} only slightly less performant.

Despite promising initial applications of \glspl{tsfm} for \gls{epf}, their performance in the \gls{epf} domain remains understudied.
This unclarity also arises from the wide variety of \glspl{tsfm} discussed in the literature, leading to questions about their comparative performance.
Furthermore, it remains unclear how \glspl{tsfm} perform relative to traditional statistical and \gls{ml} models in the context of \gls{epf}.

How effective are these so-called universal forecasters~\cite{woo_unified_2024} in the domain of \gls{epf}? This paper seeks to answer this question.
We propose a comprehensive benchmark of the state of the art \glspl{tsfm}, including variations in model sizes, as some models come in versions with different numbers of parameters.
Additionally, we include statistical and \gls{ml} models in the benchmark to allow a comparison not only between \glspl{tsfm} but also with more traditional forecasting modeling approaches.
Our evaluation uses commonly used metrics, such as \gls{rmse}, \gls{mae}, and \gls{smape}, ensuring a robust assessment of forecast performance.
Furthermore, we evaluate whether the forecast errors differ statistically using the widely applied \gls{dm} test~\cite{diebold_comparing_1995}.
To provide timely and geographically diverse results, our test interval covers the year 2024, and contains \gls{daa} prices from the largest European power markets by traded volume: Germany, France, the Netherlands, Austria, and Belgium.

% Our benchmark reveals that zero-shot \gls{epf} using \glspl{tsfm} performs comparably to statistical and \gls{ml} models across various metrics.
% We observe major performance differences among the tested \glspl{tsfm}, where Chronos-Bolt~\cite{ansari_chronos_2024} and TimeMoE~\cite{shi_time-moe_2024} exhibit some of the lowest error metrics.
% Another observation is that model performance does not always correlate with larger parameter sizes in \glspl{tsfm}. 
% When comparing \glspl{tsfm} to statistical and \gls{ml} models, none of the \glspl{tsfm} statistically outperforms a MSTL model~\cite{bandara_mstl_2021} incorporating both weekly and daily seasonality. 

We structure the remainder of this paper as follows.  
Section~\ref{sec:methodology} outlines our research approach, detailing the models we test and the evaluation metrics we use.  
Section~\ref{sec:case_study} describes the time series data, focusing on electricity prices, and specifies the parameterizations of the model.  
Section~\ref{sec:results_discussion} presents and discusses our results, including our benchmark.  
Finally, Section~\ref{sec:conclusion} concludes the paper.
% one paragraph on EPF
% problem: we do not know how amazing llms are
% therefore we do a benchmark
% compare statistcal vs llms in different european markets

%\input{03_background}
% lit review on EPF
% one paragraph traditional
% new model architectures: general intro to foundational ts models
% forecasting with LLMs in energy (we broaden to energy as gap in epf)
% already introduce the models? (-> or in scenarios?)

\section{Research Approach}\label{sec:methodology}

This section outlines our experiment setup for benchmarking.  
We test four categories of models: baseline, statistical, and \gls{ml} models, introduced in Subsection~\ref{sec:stat_ml_models_overview}, as well as a diverse set of \glspl{tsfm}, detailed in Subsection~\ref{sec:tsfm_overview}. 
We test all models in the same manner, adhering to the market schedule of the \gls{daa}, whose gate closure time is every day at noon CET~\cite{european_commission_consolidated_2021} for the next day.
To align with this market schedule, we conduct experiments daily, with each model forecasting prices for the following day, meaning that the models generate 24 price forecasts.
Baseline models and \glspl{tsfm} forecast prices directly using historical data without requiring training.
In contrast, statistical and \gls{ml} models require training prior to forecasting.
To ensure these models incorporate the latest data, we retrain them daily using updated historical price data.  
Finally, we outline the error metrics and statistical tests used to evaluate and compare model performances in Subsection~\ref{sec:error_metrics}.

\subsection{Baseline, statistical, and \acrlong{ml} models}\label{sec:stat_ml_models_overview}

We select a diverse set of time series forecasting models as reference, including three baseline models, three statistical models, and three \gls{ml} models. 
Table~\ref{tab:blsfmlf_overview} summarizes the selected models.
All models, except the baseline models, require fitting to the underlying data.
\begin{table}[h]
    \centering
    \caption{Overview of baseline, statistical, and \gls{ml} models.}
    \label{tab:blsfmlf_overview}
    \begin{tabular}{lccc}
        \toprule
        Type & Name & Library & Ref. \\
        \midrule
        \multirow{3}{*}{Baseline}& Naive& \multirow{3}{*}{StatsForecast 2.0.0~\cite{garza_statsforecast_2022}} & \multirow{3}{*}{-} \\
        & SeasonalNaiveDay    &&\\
        & SeasonalNaiveWeek   &&\\
        \addlinespace[0.1cm]
        \multirow{3}{*}{Statistical}& MSTL & \multirow{3}{*}{StatsForecast 2.0.0~\cite{garza_statsforecast_2022}} & \cite{bandara_mstl_2021} \\
        & TBATS&& \cite{de_livera_forecasting_2011} \\
        & MFLES&& \cite{tyler_mfles_2024} \\
        \addlinespace[0.1cm]
        \multirow{3}{*}{\acrshort{ml}}& ElasticNet  & \multirow{3}{*}{scikit-learn 1.6.0~\cite{pedregosa_scikit-learn_2011}} & \cite{james_introduction_2013} \\
        & KNNRegressor&& \cite{james_introduction_2013} \\
        & SVR  && \cite{vapnik_statistical_1998} \\
        \bottomrule
    \end{tabular}
\end{table}

For the baseline and statistical models, we use the Python library StatsForecast~\cite{garza_statsforecast_2022}, while we use scikit-learn~\cite{pedregosa_scikit-learn_2011} for the \gls{ml} models.
We apply each model in its default settings.
The baseline models include Naive, SeasonalNaiveDay, and SeasonalNaiveWeek, which forecast using the last known value, the value from the same time on the previous day, and the value from the same time one week earlier, respectively.

We select statistical models based on their ability to capture multiple seasonalities. Specifically, we choose MSTL, TBATS, and MFLES.
To simplify parameterization, we leverage the "Auto" functionality for TBATS and MFLES, referred to as "AutoTBATS" and "AutoMFLES"~\cite{garza_statsforecast_2022}.

For the \gls{ml} models, we test ElasticNet, KNNRegressor, and SVR, representing a regularized linear model, a nonparametric model, and a nonlinear model, respectively, to ensure a diverse comparison.
We fit the ElasticNet model using a cross-validation procedure, denoted "ElasticNetCV"~\cite{pedregosa_scikit-learn_2011}.

\subsection{\Acrlongpl{tsfm}}\label{sec:tsfm_overview}
We select a set of popular state-of-the-art \glspl{tsfm} (see Table~\ref{tab:tsfm_overview}) for our benchmark, focusing on models capable of zero-shot forecasting.
These models can perform forecasts without additional training, with pre-trained weights typically available for download.
The selected models vary in complexity, measured by the number of parameters. 
Note that the number of activated parameters in Time-MoE is significantly smaller than the total number of parameters, due to the sparse mixture of experts (MoE) architecture~\cite{shi_time-moe_2024}.
To ensure reproducibility, we provide the release dates of the model weights. 
A notable exception is the closed-source TimeGPT model, accessible only via API, with its parameter count undisclosed~\cite{garza_timegpt-1_2024}.
For TimeGPT, we document the date of access to the API instead of the weight release dates.
\begin{table}[h]
    \centering
    \begin{threeparttable}
    \caption{Overview of \glspl{tsfm}.}
    \label{tab:tsfm_overview}
    \begin{tabular}{lcccc}
        \toprule
        Model & Version(s) & Params [M] & Release & Ref. \\
        \midrule
        \multirow{2}{*}{Chronos Bolt} & Tiny, Mini, & 9, 21, & \multirow{2}{*}{ Nov 26, 2024} & \multirow{2}{*}{\cite{ansari_chronos_2024}}\\
         &  Small, Base &48, 205 & &  \\
        \addlinespace[0.1cm]
        \multirow{2}{*}{Chronos T5} & Tiny, Small,  & 8, 20,  & \multirow{2}{*}{Mar 13, 2024} &\multirow{2}{*}{\cite{ansari_chronos_2024}}\\
        & Base, Large & 46, 200 & &  \\
        \addlinespace[0.1cm]
        \multirow{2}{*}{Morai} & Small, Base,  & 14, 91,  &\multirow{2}{*}{Jun 17, 2024} &\multirow{2}{*}{\cite{woo_unified_2024}}  \\
        & Large & 311 &  & \\
        \addlinespace[0.1cm]
        TimesFM & 200M, 500M & 200, 500 & Dec 24, 2024 & \cite{das_decoder-only_2024} \\
        \addlinespace[0.1cm]
        TimeGPT & timegpt-1 & unknown & Jan 9, 2025${}^\dagger$ & \cite{garza_timegpt-1_2024} \\
        \addlinespace[0.1cm]
        TimeMoE & 50M, 200M & 50${}^{\dagger\dagger}$, 200${}^{\dagger\dagger}$ & Sep 21, 2024 & \cite{shi_time-moe_2024} \\
        \bottomrule
    \end{tabular}
    \begin{tablenotes}
    \item[$\dagger$] Date of API calls, model weights are not public.
    \item[$\dagger\dagger$] Number of activated parameters, the total number of parameters is 113M and 453M respectively.
    \end{tablenotes}
    \end{threeparttable}
\end{table}

\subsection{Forecasting performance evaluation}\label{sec:error_metrics}

We evaluate the forecasting performance of the models using widely adopted error metrics and perform \gls{dm} tests to statistically compare model performances.

Regarding error metrics, we use three widely adopted metrics in time series forecasting: \gls{rmse}, see Eq.~\eqref{eq:rmse}; \gls{mae}, see Eq.~\eqref{eq:mae}; and \gls{smape}, see Eq.~\eqref{eq:smape}.
Here, \(y_i\) denotes the true value, and \(\hat{y}_i\) represents its forecast.
\begin{align}
    \text{\acrshort{rmse}} &= \sqrt{\frac{1}{n} \sum_{i=1}^{n} \left( y_i - \hat{y}_i \right)^2}\label{eq:rmse}\\
    \text{\acrshort{mae}} &= \frac{1}{n} \sum_{i=1}^{n} \left| y_i - \hat{y}_i \right|\label{eq:mae}\\
    \text{\acrshort{smape}} &= \frac{100\%}{n} \sum_{i=1}^{n} \frac{\left| y_i - \hat{y}_i \right|}{\left| y_i \right| + \left| \hat{y}_i \right|}\label{eq:smape}
\end{align}
The \gls{rmse} emphasizes the mean of the errors, making it more sensitive to outliers compared to \gls{mae}, which focuses on the median of the errors. However, both metrics are scale-dependent and do not account for the magnitude of the true values. 
The \gls{smape} addresses this limitation by normalizing each absolute error relative to the average of the actual and predicted values. This normalization improves robustness against zero values in the true value \(y_i\), which could otherwise cause instability.

The error metrics \gls{mae}, \gls{rmse}, and \gls{smape} are used to rank models. 
However, to assess whether one model statistically outperforms another, we apply the one-sided \gls{dm} test~\cite{diebold_comparing_1995}, using the multivariate modification proposed by~\cite{ziel_day-ahead_2018}.
Traditionally, 24 independent tests are performed—one for each hour of the day~\cite{weron_electricity_2014}, with separate daily error series for each hour of the day.
This modification, however, aggregates daily errors into a single daily errors, reducing the number of required tests from 24 to 1.
Specifically, the method aggregates the 24 hourly errors using the 1-norm (sum of absolute values) to compute a single daily error metric, which serves as input for the \gls{dm} test. 
We use a maximal p-value of 0.1 and reject the null hypothesis if the test statistic falls below the critical value.
The null hypothesis states that one model's forecasts are not significantly better than the other's.
The alternative hypothesis specifies that one model has significantly superior predictive performance~\cite{diebold_comparing_1995}.% research approach?
% forecasting task

% data: add eda mean, median, std, percentiles

% statistical models: list of models and motivation for selection

% llm models: list of models and motivation for selection

% evaluation: formulas for error metrics

\section{Data and parametrization}\label{sec:case_study}

This section presents the data we use for our benchmark and the data transformations we apply in Subsection~\ref{sec:data}.  
We then outline the model parameterizations in Subsection~\ref{sec:parametrization}.  

\subsection{Data}~\label{sec:data}
We use publicly available \gls{daa} market data from multiple European countries, including Germany, France, the Netherlands, Austria, and Belgium, sourced from the ENTSO-E Transparency Platform\footnote{\url{https://transparency.entsoe.eu/}} for 2024 and parts of 2023.
In particular, the German data set also includes Luxembourg, as both belong to the same bidding zone~\cite{entso-e_bidding_2021}.
Although the testing period for our benchmark is limited to 2024, we incorporate data from the end of 2023 for model training and as input for forecasts at the start of 2024.
European \gls{dst} causes the \gls{daa} to clear for 23 hours (at the start of \gls{dst}) or 25 hours (at its end) instead of the standard 24 hours.
To simplify processing, we transform these days into 24 hour days: at the start of \gls{dst}, we interpolate to insert an additional hour, while at the end of \gls{dst}, we average to remove an hour.
This preprocessing ensures that all days in the data set have a consistent 24-hour format, facilitating compatibility with statistical and \gls{ml} models.

Although baselines, statistical methods, and \glspl{tsfm} use raw data, we preprocess the data for \gls{ml} models to enhance performance~\cite{uniejewski_variance_2018}. 
Specifically, we apply a quantile transformation, a robust technique that normalizes data by mapping them to a normal distribution based on quantiles derived from the training dataset.

\subsection{Parametrization}~\label{sec:parametrization}

In this section, we describe the model parameters used in our experiments.
The baseline models do not require parameter configuration.
For statistical and \gls{ml} models, we use 12 weeks of historical data for training, with an input size of 1 week.
This means that once trained, these models generate forecasts based on prices from the week immediately preceding the forecast period.

We configure all statistical models to account for daily and weekly seasonality.
For the MFLES and ElasticNet models, we also set seven one-day windows for \gls{cv}, corresponding to a one-week \gls{cv} period necessary for model fitting.

The \glspl{tsfm} use an input size of one week, consistent with the statistical and \gls{ml} models.

\section{Results and Discussion}\label{sec:results_discussion}
We present our benchmarking results as follows: Subsection~\ref{sec:errror_metric_results} details the error metrics, Subsection~\ref{sec:dm_test_results} reports the \gls{dm} test results, and Subsection~\ref{sec:discussion} concludes with a discussion of our results. 

\subsection{Forecasting error metric results}\label{sec:errror_metric_results}
Table~\ref{tab:benchmark} presents the benchmarking results, evaluating the forecasts of all models.
We assess each model for every country: Austria (AT), Belgium (BE), Germany (DE), France (FR), and the Netherlands (NL).
For each combination, we report three error metrics: \gls{mae}, \gls{rmse}, and \gls{smape}.
In each row, the \underline{\textbf{bold underlined}} value represents the smallest error, the \textbf{bold} value denotes the second smallest, and the \underline{underlined} value indicates the third smallest.
\begingroup
\setlength{\tabcolsep}{2pt} % Reduce horizontal padding
\setlength{\belowrulesep}{1pt} % Space below midrule
\setlength{\aboverulesep}{1pt} % Space above midrule
\begin{table*}[ht!]
    \centering
    \tiny
    \caption{Performance evaluation of forecasting models using \gls{mae}, \gls{rmse}, and \gls{smape} metrics across models and countries.}
    \label{tab:benchmark}
    %\scriptsize
    %\footnotesize
    \begin{tabular}{llcccccccccccccccccccccccccc}
\toprule
 &  & \rotatebox{90}{Naive} & \rotatebox{90}{SeasonalNaiveDay} & \rotatebox{90}{SeasonalNaiveWeek} & \rotatebox{90}{MSTL} & \rotatebox{90}{TBATS} & \rotatebox{90}{MFLES} & \rotatebox{90}{ElasticNet} & \rotatebox{90}{KNNRegressor} & \rotatebox{90}{SVR} & \rotatebox{90}{Chronos Bolt (Tiny)} & \rotatebox{90}{Chronos Bolt (Mini)} & \rotatebox{90}{Chronos Bolt (Small)} & \rotatebox{90}{Chronos Bolt (Base)} & \rotatebox{90}{Chronos T5 (Tiny)} & \rotatebox{90}{Chronos T5 (Mini)} & \rotatebox{90}{Chronos T5 (Small)} & \rotatebox{90}{Chronos T5 (Base)} & \rotatebox{90}{Chronos T5 (Large)} & \rotatebox{90}{Moirai (Small)} & \rotatebox{90}{Moirai (Base)} & \rotatebox{90}{Moirai (Large)} & \rotatebox{90}{TimesFM (200M)} & \rotatebox{90}{TimesFM (500M)} & \rotatebox{90}{TimeGPT} & \rotatebox{90}{TimeMoE (50M)} & \rotatebox{90}{TimeMoE (200M)} \\
Metric & Country &  &  &  &  &  &  &  &  &  &  &  &  &  &  &  &  &  &  &  &  &  &  &  &  &  &  \\

\midrule\multirow[t]{5}{*}{MAE} & AT & \cellcolor[rgb]{0.05,0.13,0.39}\textcolor{white}{26.35} & \cellcolor[rgb]{0.14,0.60,0.76}\textcolor{black}{22.92} & \cellcolor[rgb]{0.08,0.16,0.46}\textcolor{white}{26.02} & \cellcolor[rgb]{1.00,1.00,0.85}\textcolor{black}{\textbf{17.52}} & \cellcolor[rgb]{0.86,0.95,0.70}\textcolor{black}{19.15} & \cellcolor[rgb]{0.73,0.89,0.71}\textcolor{black}{19.99} & \cellcolor[rgb]{0.95,0.98,0.74}\textcolor{black}{18.29} & \cellcolor[rgb]{0.43,0.78,0.74}\textcolor{black}{21.18} & \cellcolor[rgb]{0.57,0.83,0.73}\textcolor{black}{20.58} & \cellcolor[rgb]{0.94,0.98,0.72}\textcolor{black}{18.44} & \cellcolor[rgb]{1.00,1.00,0.85}\textcolor{black}{\textbf{\underline{17.48}}} & \cellcolor[rgb]{1.00,1.00,0.84}\textcolor{black}{17.56} & \cellcolor[rgb]{1.00,1.00,0.85}\textcolor{black}{\underline{17.55}} & \cellcolor[rgb]{0.07,0.15,0.43}\textcolor{white}{26.12} & \cellcolor[rgb]{0.12,0.54,0.74}\textcolor{white}{23.33} & \cellcolor[rgb]{0.13,0.43,0.69}\textcolor{white}{23.97} & \cellcolor[rgb]{0.12,0.53,0.74}\textcolor{white}{23.35} & \cellcolor[rgb]{0.33,0.74,0.76}\textcolor{black}{21.64} & \cellcolor[rgb]{0.61,0.85,0.72}\textcolor{black}{20.46} & \cellcolor[rgb]{0.91,0.96,0.70}\textcolor{black}{18.82} & \cellcolor[rgb]{0.83,0.93,0.70}\textcolor{black}{19.38} & \cellcolor[rgb]{0.87,0.95,0.70}\textcolor{black}{19.06} & \cellcolor[rgb]{0.03,0.11,0.35}\textcolor{white}{26.57} & \cellcolor[rgb]{0.49,0.80,0.73}\textcolor{black}{20.92} & \cellcolor[rgb]{0.97,0.99,0.79}\textcolor{black}{17.96} & \cellcolor[rgb]{0.97,0.99,0.78}\textcolor{black}{18.04} \\
 & BE & \cellcolor[rgb]{0.12,0.18,0.53}\textcolor{white}{27.15} & \cellcolor[rgb]{0.22,0.68,0.76}\textcolor{black}{23.20} & \cellcolor[rgb]{0.03,0.11,0.35}\textcolor{white}{28.22} & \cellcolor[rgb]{1.00,1.00,0.85}\textcolor{black}{\textbf{\underline{17.43}}} & \cellcolor[rgb]{0.91,0.96,0.70}\textcolor{black}{18.99} & \cellcolor[rgb]{0.69,0.88,0.71}\textcolor{black}{20.56} & \cellcolor[rgb]{0.96,0.99,0.77}\textcolor{black}{18.11} & \cellcolor[rgb]{0.25,0.71,0.77}\textcolor{black}{22.84} & \cellcolor[rgb]{0.57,0.83,0.73}\textcolor{black}{21.12} & \cellcolor[rgb]{0.91,0.97,0.70}\textcolor{black}{18.94} & \cellcolor[rgb]{0.97,0.99,0.78}\textcolor{black}{\underline{18.05}} & \cellcolor[rgb]{0.96,0.99,0.77}\textcolor{black}{18.10} & \cellcolor[rgb]{0.96,0.98,0.75}\textcolor{black}{18.27} & \cellcolor[rgb]{0.14,0.24,0.60}\textcolor{white}{26.57} & \cellcolor[rgb]{0.22,0.68,0.76}\textcolor{black}{23.16} & \cellcolor[rgb]{0.15,0.61,0.76}\textcolor{black}{23.82} & \cellcolor[rgb]{0.28,0.72,0.76}\textcolor{black}{22.69} & \cellcolor[rgb]{0.18,0.64,0.76}\textcolor{black}{23.50} & \cellcolor[rgb]{0.78,0.92,0.71}\textcolor{black}{20.12} & \cellcolor[rgb]{0.91,0.96,0.70}\textcolor{black}{19.01} & \cellcolor[rgb]{0.82,0.93,0.70}\textcolor{black}{19.77} & \cellcolor[rgb]{0.90,0.96,0.70}\textcolor{black}{19.09} & \cellcolor[rgb]{0.13,0.45,0.69}\textcolor{white}{25.00} & \cellcolor[rgb]{0.78,0.92,0.71}\textcolor{black}{20.11} & \cellcolor[rgb]{0.98,0.99,0.81}\textcolor{black}{\textbf{17.80}} & \cellcolor[rgb]{0.95,0.98,0.74}\textcolor{black}{18.40} \\
 & DE & \cellcolor[rgb]{0.13,0.45,0.70}\textcolor{white}{29.16} & \cellcolor[rgb]{0.16,0.61,0.76}\textcolor{black}{27.82} & \cellcolor[rgb]{0.03,0.11,0.35}\textcolor{white}{32.81} & \cellcolor[rgb]{1.00,1.00,0.85}\textcolor{black}{\textbf{\underline{20.69}}} & \cellcolor[rgb]{0.86,0.94,0.70}\textcolor{black}{22.95} & \cellcolor[rgb]{0.76,0.91,0.71}\textcolor{black}{23.86} & \cellcolor[rgb]{0.98,0.99,0.80}\textcolor{black}{\textbf{21.23}} & \cellcolor[rgb]{0.29,0.73,0.76}\textcolor{black}{26.52} & \cellcolor[rgb]{0.58,0.84,0.73}\textcolor{black}{24.80} & \cellcolor[rgb]{0.93,0.97,0.70}\textcolor{black}{22.19} & \cellcolor[rgb]{0.96,0.98,0.76}\textcolor{black}{21.61} & \cellcolor[rgb]{0.96,0.99,0.77}\textcolor{black}{\underline{21.52}} & \cellcolor[rgb]{0.96,0.98,0.75}\textcolor{black}{21.65} & \cellcolor[rgb]{0.14,0.20,0.57}\textcolor{white}{31.34} & \cellcolor[rgb]{0.23,0.69,0.77}\textcolor{black}{27.02} & \cellcolor[rgb]{0.12,0.46,0.70}\textcolor{white}{29.08} & \cellcolor[rgb]{0.26,0.72,0.77}\textcolor{black}{26.74} & \cellcolor[rgb]{0.16,0.62,0.76}\textcolor{black}{27.72} & \cellcolor[rgb]{0.76,0.91,0.71}\textcolor{black}{23.84} & \cellcolor[rgb]{0.89,0.96,0.70}\textcolor{black}{22.67} & \cellcolor[rgb]{0.83,0.93,0.70}\textcolor{black}{23.20} & \cellcolor[rgb]{0.91,0.97,0.70}\textcolor{black}{22.36} & \cellcolor[rgb]{0.14,0.25,0.60}\textcolor{white}{30.88} & \cellcolor[rgb]{0.79,0.92,0.70}\textcolor{black}{23.62} & \cellcolor[rgb]{0.92,0.97,0.69}\textcolor{black}{22.32} & \cellcolor[rgb]{0.96,0.98,0.76}\textcolor{black}{21.61} \\
 & FR & \cellcolor[rgb]{0.10,0.17,0.49}\textcolor{white}{26.89} & \cellcolor[rgb]{0.51,0.81,0.73}\textcolor{black}{20.36} & \cellcolor[rgb]{0.03,0.11,0.35}\textcolor{white}{27.80} & \cellcolor[rgb]{0.98,0.99,0.80}\textcolor{black}{16.50} & \cellcolor[rgb]{0.94,0.98,0.73}\textcolor{black}{17.19} & \cellcolor[rgb]{0.90,0.96,0.70}\textcolor{black}{17.79} & \cellcolor[rgb]{0.95,0.98,0.73}\textcolor{black}{17.13} & \cellcolor[rgb]{0.30,0.73,0.76}\textcolor{black}{21.65} & \cellcolor[rgb]{0.51,0.81,0.73}\textcolor{black}{20.36} & \cellcolor[rgb]{0.97,0.99,0.78}\textcolor{black}{16.73} & \cellcolor[rgb]{1.00,1.00,0.84}\textcolor{black}{\underline{16.15}} & \cellcolor[rgb]{1.00,1.00,0.85}\textcolor{black}{\textbf{\underline{16.02}}} & \cellcolor[rgb]{1.00,1.00,0.85}\textcolor{black}{\textbf{16.03}} & \cellcolor[rgb]{0.18,0.63,0.76}\textcolor{black}{22.70} & \cellcolor[rgb]{0.27,0.72,0.77}\textcolor{black}{21.78} & \cellcolor[rgb]{0.48,0.80,0.74}\textcolor{black}{20.54} & \cellcolor[rgb]{0.34,0.75,0.76}\textcolor{black}{21.38} & \cellcolor[rgb]{0.31,0.73,0.76}\textcolor{black}{21.55} & \cellcolor[rgb]{0.76,0.91,0.71}\textcolor{black}{19.07} & \cellcolor[rgb]{0.92,0.97,0.69}\textcolor{black}{17.59} & \cellcolor[rgb]{0.85,0.94,0.70}\textcolor{black}{18.25} & \cellcolor[rgb]{0.94,0.98,0.72}\textcolor{black}{17.25} & \cellcolor[rgb]{0.33,0.74,0.76}\textcolor{black}{21.41} & \cellcolor[rgb]{0.87,0.95,0.70}\textcolor{black}{18.05} & \cellcolor[rgb]{0.97,0.99,0.78}\textcolor{black}{16.69} & \cellcolor[rgb]{0.95,0.98,0.73}\textcolor{black}{17.14} \\
 & NL & \cellcolor[rgb]{0.08,0.16,0.46}\textcolor{white}{29.31} & \cellcolor[rgb]{0.11,0.56,0.75}\textcolor{white}{26.19} & \cellcolor[rgb]{0.03,0.11,0.35}\textcolor{white}{29.95} & \cellcolor[rgb]{1.00,1.00,0.85}\textcolor{black}{\textbf{\underline{19.85}}} & \cellcolor[rgb]{0.79,0.92,0.70}\textcolor{black}{22.27} & \cellcolor[rgb]{0.56,0.83,0.73}\textcolor{black}{23.37} & \cellcolor[rgb]{0.98,0.99,0.82}\textcolor{black}{\underline{20.16}} & \cellcolor[rgb]{0.39,0.76,0.75}\textcolor{black}{24.20} & \cellcolor[rgb]{0.69,0.88,0.71}\textcolor{black}{22.79} & \cellcolor[rgb]{0.94,0.98,0.72}\textcolor{black}{20.94} & \cellcolor[rgb]{0.98,0.99,0.82}\textcolor{black}{\textbf{20.14}} & \cellcolor[rgb]{0.97,0.99,0.79}\textcolor{black}{20.39} & \cellcolor[rgb]{0.97,0.99,0.79}\textcolor{black}{20.35} & \cellcolor[rgb]{0.13,0.35,0.65}\textcolor{white}{27.56} & \cellcolor[rgb]{0.21,0.67,0.76}\textcolor{black}{25.29} & \cellcolor[rgb]{0.12,0.46,0.70}\textcolor{white}{26.80} & \cellcolor[rgb]{0.13,0.36,0.65}\textcolor{white}{27.48} & \cellcolor[rgb]{0.15,0.61,0.76}\textcolor{black}{25.83} & \cellcolor[rgb]{0.75,0.90,0.71}\textcolor{black}{22.53} & \cellcolor[rgb]{0.94,0.98,0.71}\textcolor{black}{21.01} & \cellcolor[rgb]{0.84,0.94,0.70}\textcolor{black}{21.88} & \cellcolor[rgb]{0.92,0.97,0.69}\textcolor{black}{21.16} & \cellcolor[rgb]{0.14,0.26,0.61}\textcolor{white}{28.28} & \cellcolor[rgb]{0.76,0.91,0.71}\textcolor{black}{22.46} & \cellcolor[rgb]{0.96,0.98,0.76}\textcolor{black}{20.58} & \cellcolor[rgb]{0.95,0.98,0.74}\textcolor{black}{20.75} \\
%\cline{1-28}
\midrule\multirow[t]{5}{*}{RMSE} & AT & \cellcolor[rgb]{0.14,0.20,0.57}\textcolor{white}{43.17} & \cellcolor[rgb]{0.19,0.64,0.76}\textcolor{black}{38.40} & \cellcolor[rgb]{0.03,0.11,0.35}\textcolor{white}{45.02} & \cellcolor[rgb]{0.96,0.98,0.75}\textcolor{black}{31.08} & \cellcolor[rgb]{0.87,0.95,0.70}\textcolor{black}{32.57} & \cellcolor[rgb]{0.52,0.81,0.73}\textcolor{black}{35.42} & \cellcolor[rgb]{0.91,0.96,0.70}\textcolor{black}{32.12} & \cellcolor[rgb]{0.43,0.78,0.74}\textcolor{black}{36.07} & \cellcolor[rgb]{0.61,0.85,0.72}\textcolor{black}{34.83} & \cellcolor[rgb]{0.94,0.98,0.72}\textcolor{black}{31.48} & \cellcolor[rgb]{0.97,0.99,0.78}\textcolor{black}{\textbf{30.73}} & \cellcolor[rgb]{0.96,0.98,0.76}\textcolor{black}{31.00} & \cellcolor[rgb]{1.00,1.00,0.85}\textcolor{black}{\textbf{\underline{29.89}}} & \cellcolor[rgb]{0.14,0.27,0.61}\textcolor{white}{42.30} & \cellcolor[rgb]{0.21,0.67,0.76}\textcolor{black}{38.09} & \cellcolor[rgb]{0.12,0.50,0.72}\textcolor{white}{39.98} & \cellcolor[rgb]{0.12,0.53,0.74}\textcolor{white}{39.65} & \cellcolor[rgb]{0.41,0.77,0.75}\textcolor{black}{36.25} & \cellcolor[rgb]{0.62,0.85,0.72}\textcolor{black}{34.77} & \cellcolor[rgb]{0.87,0.95,0.70}\textcolor{black}{32.52} & \cellcolor[rgb]{0.81,0.92,0.70}\textcolor{black}{33.34} & \cellcolor[rgb]{0.87,0.95,0.70}\textcolor{black}{32.55} & \cellcolor[rgb]{0.12,0.51,0.73}\textcolor{white}{39.86} & \cellcolor[rgb]{0.60,0.84,0.72}\textcolor{black}{34.90} & \cellcolor[rgb]{0.84,0.94,0.70}\textcolor{black}{32.90} & \cellcolor[rgb]{0.96,0.98,0.76}\textcolor{black}{\underline{30.96}} \\
 & BE & \cellcolor[rgb]{0.14,0.23,0.59}\textcolor{white}{38.74} & \cellcolor[rgb]{0.25,0.71,0.77}\textcolor{black}{33.24} & \cellcolor[rgb]{0.03,0.11,0.35}\textcolor{white}{40.99} & \cellcolor[rgb]{1.00,1.00,0.85}\textcolor{black}{\textbf{\underline{25.34}}} & \cellcolor[rgb]{0.94,0.98,0.71}\textcolor{black}{27.07} & \cellcolor[rgb]{0.76,0.91,0.71}\textcolor{black}{29.38} & \cellcolor[rgb]{0.96,0.98,0.75}\textcolor{black}{26.56} & \cellcolor[rgb]{0.27,0.72,0.77}\textcolor{black}{32.99} & \cellcolor[rgb]{0.57,0.83,0.73}\textcolor{black}{30.75} & \cellcolor[rgb]{0.93,0.97,0.70}\textcolor{black}{27.23} & \cellcolor[rgb]{0.98,0.99,0.80}\textcolor{black}{\textbf{25.99}} & \cellcolor[rgb]{0.96,0.99,0.77}\textcolor{black}{26.38} & \cellcolor[rgb]{0.96,0.98,0.75}\textcolor{black}{26.57} & \cellcolor[rgb]{0.14,0.23,0.59}\textcolor{white}{38.72} & \cellcolor[rgb]{0.23,0.68,0.77}\textcolor{black}{33.56} & \cellcolor[rgb]{0.12,0.55,0.75}\textcolor{white}{35.28} & \cellcolor[rgb]{0.28,0.72,0.76}\textcolor{black}{32.97} & \cellcolor[rgb]{0.17,0.63,0.76}\textcolor{black}{34.27} & \cellcolor[rgb]{0.80,0.92,0.70}\textcolor{black}{28.98} & \cellcolor[rgb]{0.89,0.96,0.70}\textcolor{black}{27.79} & \cellcolor[rgb]{0.82,0.93,0.70}\textcolor{black}{28.72} & \cellcolor[rgb]{0.90,0.96,0.70}\textcolor{black}{27.68} & \cellcolor[rgb]{0.12,0.57,0.75}\textcolor{white}{35.00} & \cellcolor[rgb]{0.79,0.92,0.71}\textcolor{black}{29.16} & \cellcolor[rgb]{0.98,0.99,0.80}\textcolor{black}{\underline{26.02}} & \cellcolor[rgb]{0.93,0.97,0.70}\textcolor{black}{27.28} \\
 & DE & \cellcolor[rgb]{0.11,0.56,0.75}\textcolor{white}{48.07} & \cellcolor[rgb]{0.38,0.76,0.75}\textcolor{black}{44.31} & \cellcolor[rgb]{0.03,0.11,0.35}\textcolor{white}{55.25} & \cellcolor[rgb]{1.00,1.00,0.85}\textcolor{black}{\textbf{35.89}} & \cellcolor[rgb]{0.93,0.97,0.69}\textcolor{black}{38.33} & \cellcolor[rgb]{0.75,0.90,0.71}\textcolor{black}{40.97} & \cellcolor[rgb]{0.97,0.99,0.79}\textcolor{black}{36.82} & \cellcolor[rgb]{0.33,0.74,0.76}\textcolor{black}{44.86} & \cellcolor[rgb]{0.66,0.87,0.72}\textcolor{black}{41.72} & \cellcolor[rgb]{0.98,0.99,0.81}\textcolor{black}{36.50} & \cellcolor[rgb]{0.99,1.00,0.84}\textcolor{black}{\underline{36.10}} & \cellcolor[rgb]{0.98,0.99,0.82}\textcolor{black}{36.40} & \cellcolor[rgb]{1.00,1.00,0.85}\textcolor{black}{\textbf{\underline{35.85}}} & \cellcolor[rgb]{0.13,0.40,0.67}\textcolor{white}{50.00} & \cellcolor[rgb]{0.33,0.74,0.76}\textcolor{black}{44.75} & \cellcolor[rgb]{0.12,0.52,0.73}\textcolor{white}{48.54} & \cellcolor[rgb]{0.40,0.77,0.75}\textcolor{black}{44.15} & \cellcolor[rgb]{0.30,0.73,0.76}\textcolor{black}{45.12} & \cellcolor[rgb]{0.90,0.96,0.70}\textcolor{black}{38.82} & \cellcolor[rgb]{0.92,0.97,0.69}\textcolor{black}{38.49} & \cellcolor[rgb]{0.89,0.96,0.70}\textcolor{black}{39.02} & \cellcolor[rgb]{0.98,0.99,0.82}\textcolor{black}{36.43} & \cellcolor[rgb]{0.23,0.69,0.77}\textcolor{black}{46.00} & \cellcolor[rgb]{0.84,0.94,0.70}\textcolor{black}{39.75} & \cellcolor[rgb]{0.86,0.94,0.70}\textcolor{black}{39.45} & \cellcolor[rgb]{0.97,0.99,0.79}\textcolor{black}{36.86} \\
 & FR & \cellcolor[rgb]{0.13,0.19,0.54}\textcolor{white}{35.44} & \cellcolor[rgb]{0.39,0.76,0.75}\textcolor{black}{28.34} & \cellcolor[rgb]{0.03,0.11,0.35}\textcolor{white}{37.09} & \cellcolor[rgb]{0.98,0.99,0.82}\textcolor{black}{22.09} & \cellcolor[rgb]{0.97,0.99,0.78}\textcolor{black}{22.58} & \cellcolor[rgb]{0.94,0.98,0.72}\textcolor{black}{23.33} & \cellcolor[rgb]{0.94,0.98,0.71}\textcolor{black}{23.37} & \cellcolor[rgb]{0.23,0.69,0.77}\textcolor{black}{29.62} & \cellcolor[rgb]{0.55,0.82,0.73}\textcolor{black}{27.11} & \cellcolor[rgb]{0.96,0.99,0.77}\textcolor{black}{22.64} & \cellcolor[rgb]{0.99,1.00,0.84}\textcolor{black}{\underline{21.86}} & \cellcolor[rgb]{1.00,1.00,0.85}\textcolor{black}{\textbf{\underline{21.66}}} & \cellcolor[rgb]{1.00,1.00,0.85}\textcolor{black}{\textbf{21.75}} & \cellcolor[rgb]{0.12,0.55,0.74}\textcolor{white}{31.54} & \cellcolor[rgb]{0.16,0.62,0.76}\textcolor{black}{30.67} & \cellcolor[rgb]{0.36,0.75,0.75}\textcolor{black}{28.54} & \cellcolor[rgb]{0.20,0.65,0.76}\textcolor{black}{30.19} & \cellcolor[rgb]{0.20,0.66,0.76}\textcolor{black}{30.11} & \cellcolor[rgb]{0.78,0.92,0.71}\textcolor{black}{25.51} & \cellcolor[rgb]{0.92,0.97,0.69}\textcolor{black}{23.66} & \cellcolor[rgb]{0.85,0.94,0.70}\textcolor{black}{24.62} & \cellcolor[rgb]{0.95,0.98,0.73}\textcolor{black}{23.13} & \cellcolor[rgb]{0.25,0.71,0.77}\textcolor{black}{29.41} & \cellcolor[rgb]{0.87,0.95,0.70}\textcolor{black}{24.38} & \cellcolor[rgb]{0.97,0.99,0.79}\textcolor{black}{22.46} & \cellcolor[rgb]{0.95,0.98,0.74}\textcolor{black}{23.10} \\
 & NL & \cellcolor[rgb]{0.14,0.20,0.57}\textcolor{white}{46.33} & \cellcolor[rgb]{0.21,0.67,0.76}\textcolor{black}{40.84} & \cellcolor[rgb]{0.03,0.11,0.35}\textcolor{white}{48.33} & \cellcolor[rgb]{1.00,1.00,0.85}\textcolor{black}{\textbf{\underline{32.18}}} & \cellcolor[rgb]{0.87,0.95,0.70}\textcolor{black}{35.04} & \cellcolor[rgb]{0.62,0.85,0.72}\textcolor{black}{37.41} & \cellcolor[rgb]{0.96,0.98,0.76}\textcolor{black}{33.43} & \cellcolor[rgb]{0.23,0.69,0.77}\textcolor{black}{40.53} & \cellcolor[rgb]{0.71,0.89,0.71}\textcolor{black}{36.76} & \cellcolor[rgb]{0.97,0.99,0.78}\textcolor{black}{33.11} & \cellcolor[rgb]{1.00,1.00,0.85}\textcolor{black}{\textbf{32.29}} & \cellcolor[rgb]{0.97,0.99,0.78}\textcolor{black}{33.08} & \cellcolor[rgb]{0.99,1.00,0.83}\textcolor{black}{\underline{32.54}} & \cellcolor[rgb]{0.12,0.50,0.72}\textcolor{white}{42.93} & \cellcolor[rgb]{0.23,0.69,0.77}\textcolor{black}{40.63} & \cellcolor[rgb]{0.22,0.68,0.76}\textcolor{black}{40.74} & \cellcolor[rgb]{0.13,0.43,0.69}\textcolor{white}{43.62} & \cellcolor[rgb]{0.11,0.56,0.75}\textcolor{white}{42.30} & \cellcolor[rgb]{0.84,0.94,0.70}\textcolor{black}{35.41} & \cellcolor[rgb]{0.95,0.98,0.73}\textcolor{black}{33.73} & \cellcolor[rgb]{0.84,0.94,0.70}\textcolor{black}{35.52} & \cellcolor[rgb]{0.96,0.98,0.75}\textcolor{black}{33.50} & \cellcolor[rgb]{0.14,0.59,0.76}\textcolor{black}{41.91} & \cellcolor[rgb]{0.80,0.92,0.70}\textcolor{black}{36.01} & \cellcolor[rgb]{0.95,0.98,0.75}\textcolor{black}{33.56} & \cellcolor[rgb]{0.94,0.98,0.73}\textcolor{black}{33.80} \\
%\cline{1-28}
\midrule\multirow[t]{5}{*}{SMAPE [\%]} & AT & \cellcolor[rgb]{0.33,0.74,0.76}\textcolor{black}{19.81} & \cellcolor[rgb]{0.15,0.60,0.76}\textcolor{black}{20.97} & \cellcolor[rgb]{0.14,0.28,0.62}\textcolor{white}{22.81} & \cellcolor[rgb]{1.00,1.00,0.85}\textcolor{black}{\textbf{\underline{15.95}}} & \cellcolor[rgb]{0.96,0.99,0.77}\textcolor{black}{16.52} & \cellcolor[rgb]{0.94,0.98,0.73}\textcolor{black}{16.79} & \cellcolor[rgb]{0.94,0.98,0.71}\textcolor{black}{16.93} & \cellcolor[rgb]{0.56,0.83,0.73}\textcolor{black}{18.91} & \cellcolor[rgb]{0.57,0.83,0.73}\textcolor{black}{18.87} & \cellcolor[rgb]{0.90,0.96,0.70}\textcolor{black}{17.21} & \cellcolor[rgb]{0.96,0.99,0.77}\textcolor{black}{16.51} & \cellcolor[rgb]{0.96,0.99,0.77}\textcolor{black}{16.54} & \cellcolor[rgb]{0.95,0.98,0.73}\textcolor{black}{16.78} & \cellcolor[rgb]{0.13,0.19,0.54}\textcolor{white}{23.46} & \cellcolor[rgb]{0.12,0.55,0.75}\textcolor{white}{21.32} & \cellcolor[rgb]{0.12,0.46,0.70}\textcolor{white}{21.77} & \cellcolor[rgb]{0.12,0.55,0.74}\textcolor{white}{21.33} & \cellcolor[rgb]{0.36,0.75,0.75}\textcolor{black}{19.72} & \cellcolor[rgb]{0.85,0.94,0.70}\textcolor{black}{17.56} & \cellcolor[rgb]{0.95,0.98,0.75}\textcolor{black}{16.66} & \cellcolor[rgb]{0.96,0.98,0.75}\textcolor{black}{16.63} & \cellcolor[rgb]{0.89,0.96,0.70}\textcolor{black}{17.33} & \cellcolor[rgb]{0.03,0.11,0.35}\textcolor{white}{24.38} & \cellcolor[rgb]{0.78,0.91,0.71}\textcolor{black}{18.08} & \cellcolor[rgb]{0.99,1.00,0.83}\textcolor{black}{\textbf{16.09}} & \cellcolor[rgb]{0.97,0.99,0.79}\textcolor{black}{\underline{16.36}} \\
 & BE & \cellcolor[rgb]{0.32,0.74,0.76}\textcolor{black}{23.87} & \cellcolor[rgb]{0.12,0.52,0.73}\textcolor{white}{25.61} & \cellcolor[rgb]{0.03,0.11,0.35}\textcolor{white}{28.79} & \cellcolor[rgb]{1.00,1.00,0.85}\textcolor{black}{\textbf{19.60}} & \cellcolor[rgb]{0.97,0.99,0.79}\textcolor{black}{\underline{20.03}} & \cellcolor[rgb]{0.94,0.98,0.72}\textcolor{black}{20.53} & \cellcolor[rgb]{0.94,0.98,0.73}\textcolor{black}{20.50} & \cellcolor[rgb]{0.30,0.73,0.76}\textcolor{black}{23.99} & \cellcolor[rgb]{0.52,0.81,0.73}\textcolor{black}{22.92} & \cellcolor[rgb]{0.84,0.94,0.70}\textcolor{black}{21.48} & \cellcolor[rgb]{0.93,0.97,0.70}\textcolor{black}{20.67} & \cellcolor[rgb]{0.92,0.97,0.69}\textcolor{black}{20.78} & \cellcolor[rgb]{0.88,0.95,0.70}\textcolor{black}{21.09} & \cellcolor[rgb]{0.12,0.18,0.52}\textcolor{white}{27.91} & \cellcolor[rgb]{0.11,0.56,0.75}\textcolor{white}{25.35} & \cellcolor[rgb]{0.13,0.44,0.69}\textcolor{white}{26.08} & \cellcolor[rgb]{0.18,0.64,0.76}\textcolor{black}{24.78} & \cellcolor[rgb]{0.12,0.53,0.74}\textcolor{white}{25.52} & \cellcolor[rgb]{0.86,0.95,0.70}\textcolor{black}{21.24} & \cellcolor[rgb]{0.93,0.97,0.70}\textcolor{black}{20.68} & \cellcolor[rgb]{0.94,0.98,0.72}\textcolor{black}{20.57} & \cellcolor[rgb]{0.93,0.97,0.70}\textcolor{black}{20.72} & \cellcolor[rgb]{0.14,0.21,0.59}\textcolor{white}{27.54} & \cellcolor[rgb]{0.89,0.96,0.70}\textcolor{black}{21.08} & \cellcolor[rgb]{1.00,1.00,0.85}\textcolor{black}{\textbf{\underline{19.57}}} & \cellcolor[rgb]{0.97,0.99,0.78}\textcolor{black}{20.12} \\
 & DE & \cellcolor[rgb]{0.49,0.80,0.73}\textcolor{black}{23.81} & \cellcolor[rgb]{0.12,0.57,0.75}\textcolor{white}{26.32} & \cellcolor[rgb]{0.09,0.16,0.47}\textcolor{white}{29.49} & \cellcolor[rgb]{1.00,1.00,0.85}\textcolor{black}{\textbf{\underline{19.92}}} & \cellcolor[rgb]{0.95,0.98,0.74}\textcolor{black}{20.80} & \cellcolor[rgb]{0.93,0.97,0.70}\textcolor{black}{21.14} & \cellcolor[rgb]{0.94,0.98,0.71}\textcolor{black}{21.06} & \cellcolor[rgb]{0.41,0.77,0.75}\textcolor{black}{24.21} & \cellcolor[rgb]{0.33,0.74,0.76}\textcolor{black}{24.66} & \cellcolor[rgb]{0.90,0.96,0.70}\textcolor{black}{21.46} & \cellcolor[rgb]{0.91,0.96,0.70}\textcolor{black}{21.44} & \cellcolor[rgb]{0.91,0.97,0.70}\textcolor{black}{21.34} & \cellcolor[rgb]{0.91,0.96,0.70}\textcolor{black}{21.43} & \cellcolor[rgb]{0.03,0.11,0.35}\textcolor{white}{30.18} & \cellcolor[rgb]{0.12,0.58,0.75}\textcolor{white}{26.23} & \cellcolor[rgb]{0.13,0.40,0.67}\textcolor{white}{27.44} & \cellcolor[rgb]{0.21,0.67,0.76}\textcolor{black}{25.43} & \cellcolor[rgb]{0.16,0.62,0.76}\textcolor{black}{25.87} & \cellcolor[rgb]{0.84,0.94,0.70}\textcolor{black}{22.02} & \cellcolor[rgb]{0.92,0.97,0.69}\textcolor{black}{21.26} & \cellcolor[rgb]{0.95,0.98,0.75}\textcolor{black}{20.77} & \cellcolor[rgb]{0.94,0.98,0.72}\textcolor{black}{20.99} & \cellcolor[rgb]{0.12,0.18,0.52}\textcolor{white}{29.19} & \cellcolor[rgb]{0.92,0.97,0.69}\textcolor{black}{21.29} & \cellcolor[rgb]{0.98,0.99,0.80}\textcolor{black}{\underline{20.39}} & \cellcolor[rgb]{0.98,0.99,0.82}\textcolor{black}{\textbf{20.23}} \\
 & FR & \cellcolor[rgb]{0.40,0.77,0.75}\textcolor{black}{29.98} & \cellcolor[rgb]{0.30,0.73,0.76}\textcolor{black}{30.56} & \cellcolor[rgb]{0.03,0.11,0.35}\textcolor{white}{36.42} & \cellcolor[rgb]{0.98,0.99,0.80}\textcolor{black}{\underline{25.61}} & \cellcolor[rgb]{1.00,1.00,0.85}\textcolor{black}{\textbf{25.20}} & \cellcolor[rgb]{0.98,0.99,0.80}\textcolor{black}{25.66} & \cellcolor[rgb]{0.93,0.97,0.70}\textcolor{black}{26.56} & \cellcolor[rgb]{0.24,0.70,0.77}\textcolor{black}{30.91} & \cellcolor[rgb]{0.49,0.80,0.73}\textcolor{black}{29.43} & \cellcolor[rgb]{0.95,0.98,0.73}\textcolor{black}{26.26} & \cellcolor[rgb]{0.98,0.99,0.80}\textcolor{black}{25.68} & \cellcolor[rgb]{0.97,0.99,0.79}\textcolor{black}{25.73} & \cellcolor[rgb]{0.97,0.99,0.79}\textcolor{black}{25.76} & \cellcolor[rgb]{0.12,0.47,0.71}\textcolor{white}{32.86} & \cellcolor[rgb]{0.16,0.62,0.76}\textcolor{black}{31.72} & \cellcolor[rgb]{0.29,0.73,0.76}\textcolor{black}{30.58} & \cellcolor[rgb]{0.13,0.45,0.70}\textcolor{white}{33.02} & \cellcolor[rgb]{0.13,0.44,0.69}\textcolor{white}{33.09} & \cellcolor[rgb]{0.91,0.97,0.70}\textcolor{black}{26.74} & \cellcolor[rgb]{0.96,0.98,0.76}\textcolor{black}{26.03} & \cellcolor[rgb]{0.95,0.98,0.74}\textcolor{black}{26.16} & \cellcolor[rgb]{0.96,0.98,0.76}\textcolor{black}{26.01} & \cellcolor[rgb]{0.16,0.62,0.76}\textcolor{black}{31.73} & \cellcolor[rgb]{0.95,0.98,0.75}\textcolor{black}{26.10} & \cellcolor[rgb]{1.00,1.00,0.85}\textcolor{black}{\textbf{\underline{25.17}}} & \cellcolor[rgb]{0.96,0.98,0.76}\textcolor{black}{25.99} \\
 & NL & \cellcolor[rgb]{0.33,0.74,0.76}\textcolor{black}{23.41} & \cellcolor[rgb]{0.12,0.46,0.70}\textcolor{white}{25.34} & \cellcolor[rgb]{0.03,0.11,0.35}\textcolor{white}{27.84} & \cellcolor[rgb]{1.00,1.00,0.85}\textcolor{black}{\textbf{\underline{19.68}}} & \cellcolor[rgb]{0.94,0.98,0.71}\textcolor{black}{20.61} & \cellcolor[rgb]{0.91,0.96,0.70}\textcolor{black}{20.84} & \cellcolor[rgb]{0.91,0.97,0.70}\textcolor{black}{20.82} & \cellcolor[rgb]{0.52,0.81,0.73}\textcolor{black}{22.67} & \cellcolor[rgb]{0.43,0.78,0.74}\textcolor{black}{23.01} & \cellcolor[rgb]{0.86,0.95,0.70}\textcolor{black}{21.15} & \cellcolor[rgb]{0.93,0.97,0.69}\textcolor{black}{20.72} & \cellcolor[rgb]{0.89,0.96,0.70}\textcolor{black}{21.02} & \cellcolor[rgb]{0.92,0.97,0.69}\textcolor{black}{20.78} & \cellcolor[rgb]{0.12,0.18,0.53}\textcolor{white}{27.01} & \cellcolor[rgb]{0.24,0.70,0.77}\textcolor{black}{23.83} & \cellcolor[rgb]{0.14,0.28,0.62}\textcolor{white}{26.35} & \cellcolor[rgb]{0.14,0.24,0.60}\textcolor{white}{26.60} & \cellcolor[rgb]{0.15,0.60,0.76}\textcolor{black}{24.55} & \cellcolor[rgb]{0.83,0.93,0.70}\textcolor{black}{21.43} & \cellcolor[rgb]{0.95,0.98,0.73}\textcolor{black}{20.47} & \cellcolor[rgb]{0.97,0.99,0.78}\textcolor{black}{20.14} & \cellcolor[rgb]{0.94,0.98,0.73}\textcolor{black}{20.50} & \cellcolor[rgb]{0.07,0.14,0.42}\textcolor{white}{27.51} & \cellcolor[rgb]{0.87,0.95,0.70}\textcolor{black}{21.09} & \cellcolor[rgb]{1.00,1.00,0.84}\textcolor{black}{\textbf{19.77}} & \cellcolor[rgb]{0.98,0.99,0.80}\textcolor{black}{\underline{20.03}} \\
%\cline{1-28}
\bottomrule
\end{tabular}

\end{table*}
\endgroup

Within each model type group, we make the following observations: For the baseline models, the SeasonalNaiveDay outperforms across countries for \gls{mae} and \gls{rmse}, while the Naive model performs best for \gls{smape}.

Among statistical models, the MSTL model consistently achieves the best results in all countries and metrics, except for the \gls{smape} in France, where TBATS yields a smaller error.

Among the \gls{ml} models, the ElasticNet model demonstrates superior performance across all countries and metrics.

Among the \glspl{tsfm}, two models stand out: Chronos Bolt and TimeMoE.
Chronos Bolt achieves strong results across countries for \gls{mae} and \gls{rmse}, while TimeMoE achieves lower \gls{smape}.
Additionally, the Chronos Bolt models significantly outperform the Chronos T5 models.
When comparing models with different numbers of parameters, no clear pattern emerges, except for the smaller models generally underperforming relative to their larger counterparts within the same type (e.g., Chronos Bolt (Tiny), Chronos T5 (Tiny), Moirai (Small)).
A notable exception is TimesFM, where performance decreases from the 200M model to the 500M model.

In general, for \gls{mae} and \gls{rmse}, the best performing model for each country is the MSTL or a Chronos Bolt model (Mini, Small, or Base).
For \gls{smape}, the top models are MSTL or TimeMoE.

\subsection{\Acrlong{dm} test results}\label{sec:dm_test_results}
\begin{figure}[h]
    \centering
    \includegraphics[width=\linewidth]{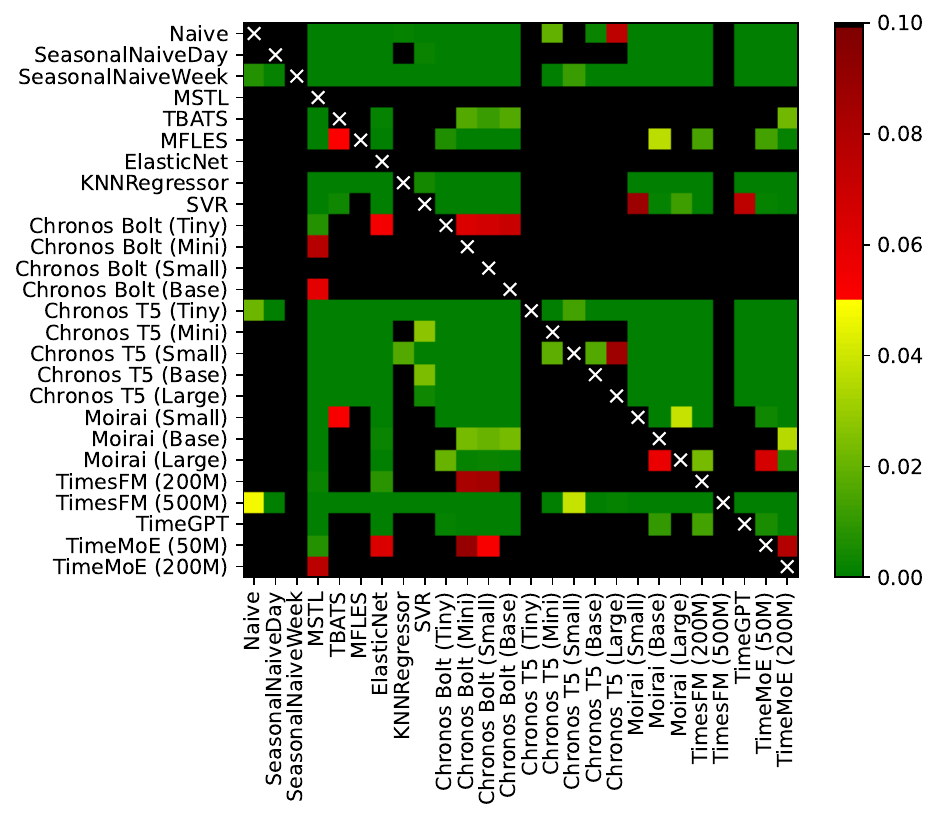}
    \caption{\Gls{dm} test results for Germany.}
    \label{fig:dm_de}
\end{figure}

Fig.~\ref{fig:dm_de} presents the \gls{dm} test results for Germany, the largest European power market by traded volume.
Due to space constraints, we provide the results for other countries in the appendix.
We present the test results as heat maps that illustrate the p-values of the \gls{dm} test.
In these heat maps, p-values above 0.1 are depicted in black, indicating acceptance of the null hypothesis—that neither model is significantly better than the other.
Lower p-values indicate cases where the forecast on the x-axis is significantly more performant than that on the y-axis.
For example, in Germany, the MSTL model statistically outperforms all Chronos T5 models, but is not significantly more performant than the ElasticNet model.
For Germany, we make several observations.
The results of the \gls{dm} test indicate that most of the models outperform the baseline models (Naive, SeasonalNaiveDay, and SeasonalNaiveWeek), with notable exceptions being Chronos Bolt (Tiny) and TimesFM (500M).
Furthermore, all Chronos T5 models fail to surpass the SeasonalNaiveDay baseline. 
Comparing \glspl{tsfm} with statistical and \gls{ml} models, none of the \glspl{tsfm} is significantly superior to the MSTL and ElasticNet models.
Among all models, MSTL surpasses the most other models in performance, with only two exceptions: ElasticNet and Chronos Bolt (Small).
 
The results in other markets follow a similar pattern.
The Chronos T5 models generally perform poorly, with only the Large variant surpassing all naive models in Austria but failing to outperform SeasonalNaiveDay in other countries.
MSTL performs well across markets, except in France, where Chronos Bolt (Small) achieves significantly higher performance. 
In both France and Austria, multiple Chronos Bolt models (Mini, Small, and Base) outperform ElasticNet.
Additionally, these Chronos Bolt models consistently demonstrate higher performance than other \glspl{tsfm}, with the exception of TimeMoE models, which they surpass only in specific countries and parameter configurations. 
The Chronos Bolt models (Mini, Small, and Base) have no statistically significant differences in performance between themselves in all countries tested. 

\subsection{Discussion}\label{sec:discussion}
The superior performance of SeasonalNaiveDay over Naive in terms of \gls{mae} and \gls{rmse} emphasizes the importance of daily seasonality in \gls{epf}.
In contrast, Naive's better performance in \gls{smape} highlights the influence of scale sensitivity, suggesting that error metrics must be carefully chosen based on forecasting objectives.

Extending the analysis to statistical and \gls{ml} models, our results indicate that the MSTL model is versatile, performing well in both error metrics and statistical tests in all countries tested.
This observation suggests that well-calibrated statistical approaches can remain competitive with, and in some cases rival, more complex \gls{ml} models.

Focusing on \glspl{tsfm}, the superior performance of Chronos Bolt over Chronos T5 in both error metrics and statistically significant performance validates the improvements introduced in the more recent Chronos Bolt model. These updates demonstrate tangible gains in forecasting performance. 

Chronos Bolt models excel in minimizing \gls{mae} and \gls{rmse} while outperforming most competitors in statistical tests, although TimeMoE achieves better performance for \gls{smape}. Despite minor variations in performance across metrics, the different Chronos Bolt variants (Mini, Small, and Base) show no statistically significant differences in performance across the countries tested. 

Model selection ultimately depends on the error metric most relevant to the application.
MSTL stands out as a reliable all-rounder, consistently achieving strong performance across all metrics and statistical tests.
Among \glspl{tsfm}, the Chronos Bolt models (Mini, Small, and Base) consistently outperform most competitors, particularly excelling in minimizing \gls{mae} and \gls{rmse}.
Similarly, TimeMoE (50M or 200M) is well-suited for otimizing \gls{smape}.
For applications requiring parameter efficiency (that is, \glspl{tsfm} with fewer parameters for faster inference), Chronos Bolt (Mini) and TimeMoE (50M) offer practical choices.
Depending on the relevant metric, Chronos Bolt excels in \gls{mae} and \gls{rmse}, while TimeMoE performs best for \gls{smape}.

Performance comparison alone does not provide a definitive recommendation, as factors like model interpretability and ease of use also influence the model selection process.
A key limitation of \glspl{tsfm} is their lack of interpretability, which may hinder their adoption by practitioners who require transparency for real-world decision making.
However, \glspl{tsfm} offer a significant advantage in ease of use, enabling inference without the need for prior training. 
Despite this benefit, the extensive variety of \glspl{tsfm} may pose a challenge for adoption, as selecting the most suitable model remains a nontrivial task.
\section{Conclusion}\label{sec:conclusion}
This paper provides a comprehensive benchmark of \glspl{tsfm}, evaluating their performance in the largest European power markets using multiple error metrics. 
Our findings reveal that some \glspl{tsfm} can compete with statistical and \gls{ml} forecasting methods, with MSTL consistently demonstrating robust performance across metrics and countries. 
However, identifying a single optimal \gls{tsfm} remains challenging, as different models excel for different error metrics (e.g., Chronos Bolt for \gls{mae} and \gls{rmse}, and TimeMoE for \gls{smape}). 
Additionally, increasing the number of parameters does not guarantee improved performance, complicating the choice of both the type and size of the model.
The lack of interpretability in \glspl{tsfm} further limits their applicability in use cases where explainability is crucial.

For noncritical applications, where ease of use and minimal training requirements outweigh performance and interpretability concerns, \glspl{tsfm} offer a practical solution.
However, for scenarios that require high performance and interpretability, statistical methods remain the most reliable choice.

Our study has several limitations, which present opportunities for future research. We did not incorporate exogenous variables into our forecasting models, although their inclusion could significantly enhance performance. Additionally, the study's geographical scope is restricted to five European countries, leaving room for analysis in other markets. 
Moreover, our evaluation relies on a single model configuration. Future research could investigate the effects of varying the input sizes for \glspl{tsfm} and extending the forecast horizon.

\section*{Acknowledgment}
During the preparation of this work the authors used ChatGPT and Writefull in order to improve readability and language. After using this tool/service, the authors reviewed and edited the content as needed and take full responsibility for the content of the publication.



\newpage
\bibliographystyle{IEEEtran}
\bibliography{99_bibliography_th, 99_bibliography_as}



% \end{thebibliography}
\appendix
The following figures present additional results, specifically the \gls{dm} test outcomes for the studied countries, excluding Germany, which we reported in the main part of the paper.
\begin{figure}[h]
    \centering
    \includegraphics[width=\linewidth]{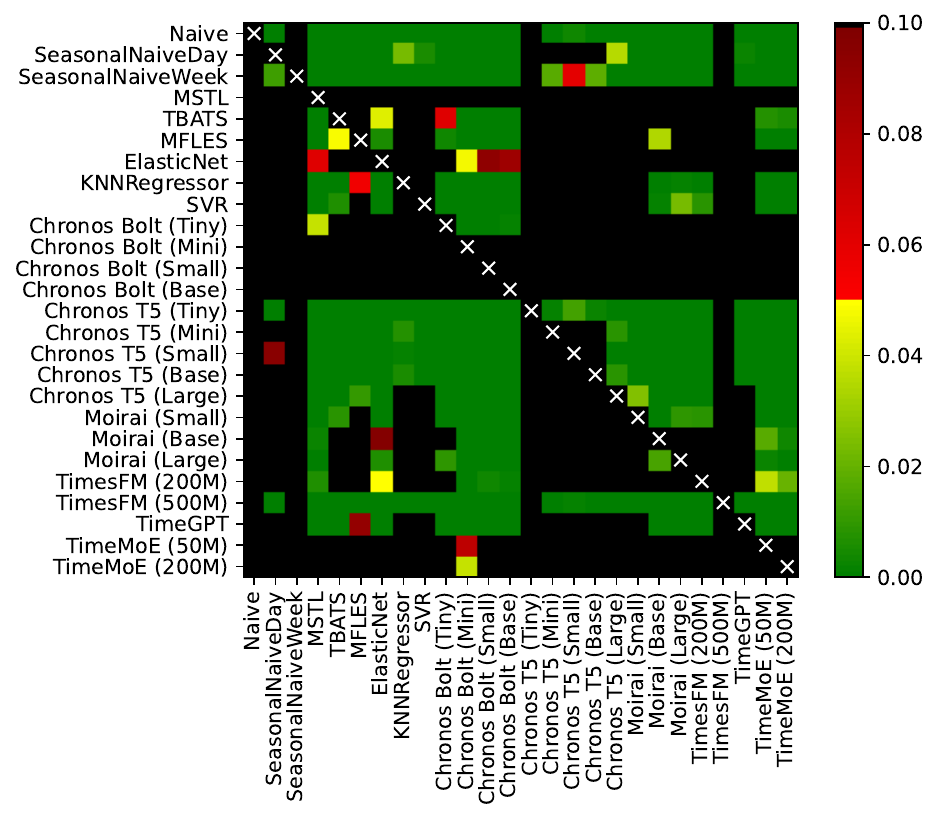}
    \caption{\Gls{dm} test results for Austria.}
    \label{fig:dm_at}
\end{figure}
\begin{figure}[h]
    \centering
    \includegraphics[width=\linewidth]{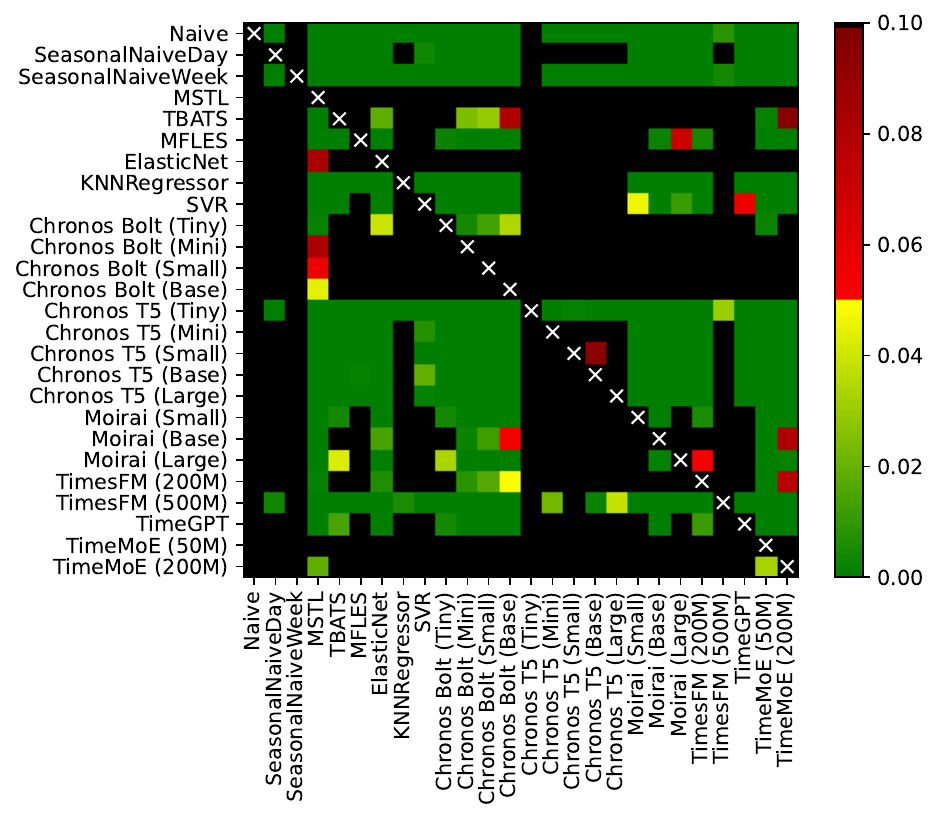}
    \caption{\Gls{dm} test results for Belgium.}
    \label{fig:dm_be}
\end{figure}
\begin{figure}[h]
    \centering
    \includegraphics[width=\linewidth]{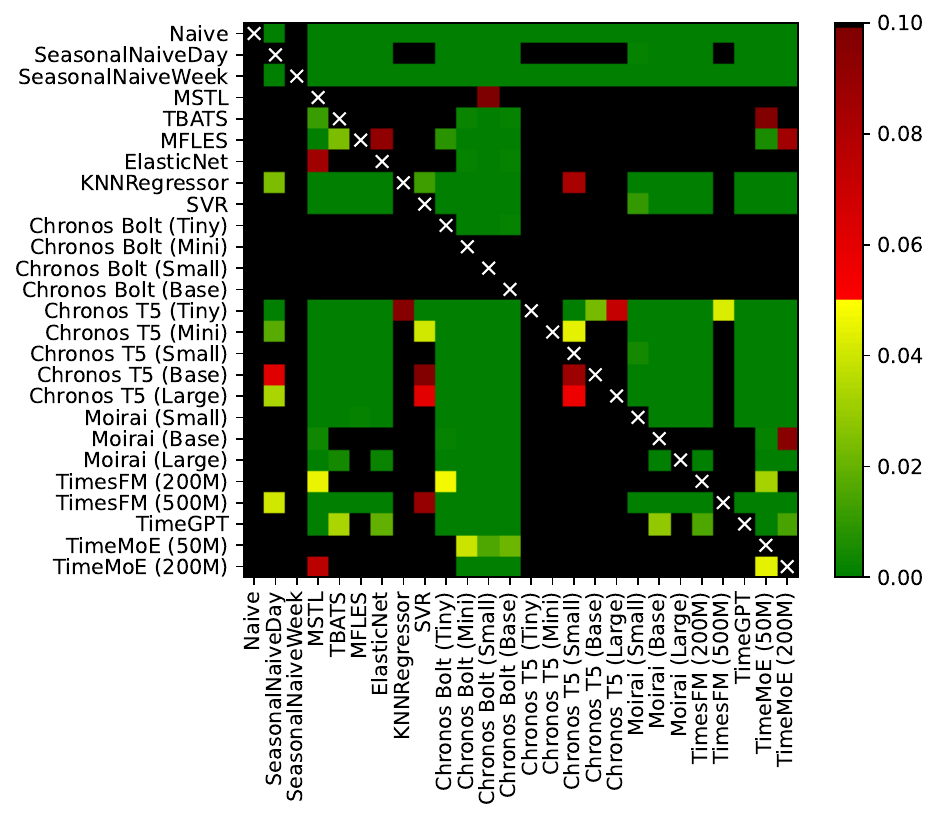}
    \caption{\Gls{dm} test results for France.}
    \label{fig:dm_fr}
\end{figure}
\begin{figure}[h]
    \centering
    \includegraphics[width=\linewidth]{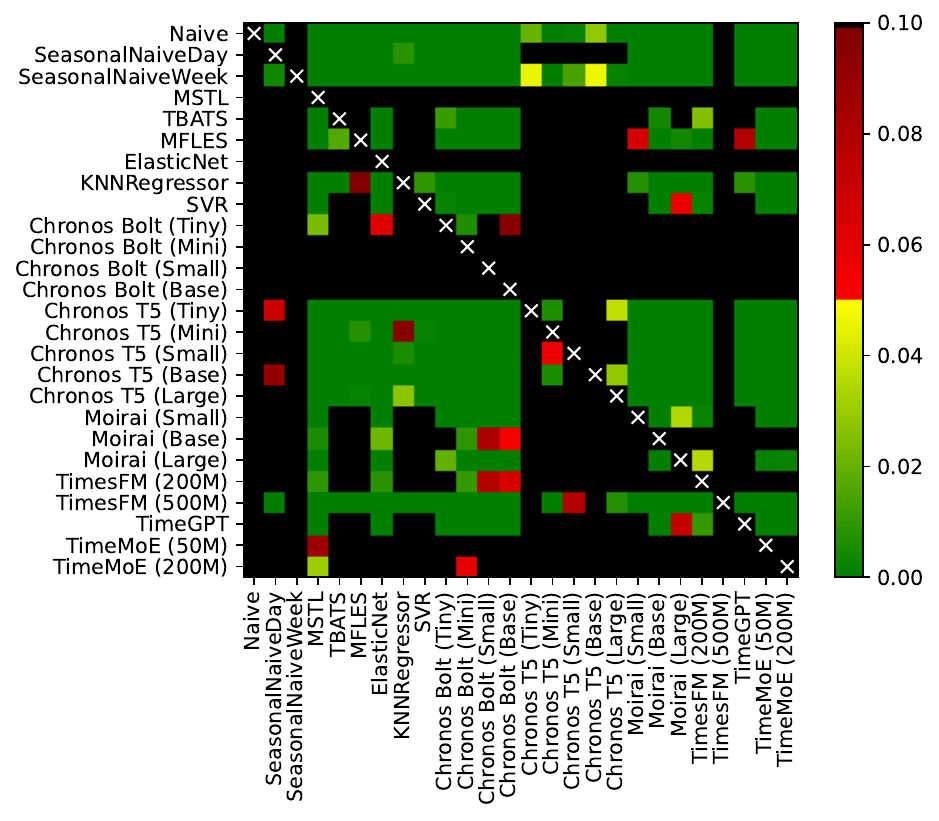}
    \caption{\Gls{dm} test results for the Netherlands.}
    \label{fig:dm_nl}
\end{figure}

\end{document}